\documentclass[sigconf]{acmart}
\usepackage{tabularx}
\usepackage{graphicx}
\usepackage{siunitx}
\usepackage{cleveref}
\usepackage{enumitem}
\usepackage{microtype}
\DeclareSIUnit{\bit}{bit}
\DeclareSIUnit{\byte}{B}
\AtBeginDocument{%
  }

\copyrightyear{2026}\acmYear{2026}
\setcopyright{cc}
\setcctype{by}
\acmConference[MobiHoc '26]{The Twenty-seventh International Symposium on Theory, Algorithmic Foundations, and Protocol Design for Mobile Networks and Mobile Computing}{November 23-26, 2026}{Tokyo, Japan}
\acmBooktitle{The Twenty-seventh International Symposium on Theory, Algorithmic Foundations, and Protocol Design for Mobile Networks and Mobile Computing (MobiHoc '26), November 23-26, 2026, Tokyo, Japan}
\acmDOI{10.1145/3842721.3850112}
\acmISBN{979-8-4007-2975-1/2026/11}

\begin{document}

\title{Efficient Secure Federated Learning via Information-Theoretically Secure Key Distribution: A Medical Imaging Case Study}

\author{Ivan Donà}
\orcid{0009-0003-5556-4380}
\affiliation{%
  \institution{University of Trento}
  \city{Trento}
  \country{Italy}
}
\affiliation{%
  \institution{Huawei Technologies Duesseldorf}
  \city{Munich}
  \country{Germany}
}

\author{Hans H. Brunner}
\correspondingauthor{}
\email{hans.brunner@huawei.com}
\orcid{0000-0002-4878-0435}
\affiliation{%
  \institution{Huawei Technologies Duesseldorf}
  \city{Munich}
  \country{Germany}
}
\author{Álvaro Troyano Olivas}
\orcid{0009-0002-1919-1225}
\affiliation{%
  \institution{Universidad Politécnica de Madrid}
  \city{Madrid}
  \country{Spain}
}
\affiliation{%
  \institution{Huawei Technologies Duesseldorf}
  \city{Munich}
  \country{Germany}
}

\author{Chi-Hang Fred Fung}
\orcid{0000-0001-6918-1047}
\affiliation{%
  \institution{Huawei Technologies Duesseldorf}
  \city{Munich}
  \country{Germany}
}

\author{Momtchil Peev}
\orcid{0000-0002-7928-0409}
\affiliation{%
  \institution{Huawei Technologies Duesseldorf}
  \city{Munich}
  \country{Germany}
}
    
\author{Giovanni Iacca}
\correspondingauthor{}
\email{giovanni.iacca@unitn.it}
\orcid{0000-0001-9723-1830}
\affiliation{%
  \institution{University of Trento}
  \city{Trento}
  \country{Italy}
}

\renewcommand{\shortauthors}{Donà et al.}

\begin{abstract}
Federated Learning (FL) enables collaborative training of models across institutions without centralizing sensitive data, making it well-suited for privacy-concerned applications, such as medical imaging. To protect FL model updates during secure aggregation, additive masking is commonly employed. However, its underlying classical key establishment is only computationally secure. On the other hand, physics-based Information-Theoretically Secure (ITS) key exchange introduces practical constraints: finite key generation rates and time-limited storage severely limit throughput and sustained training of uncompressed models. In this work, we address this bottleneck by developing an FL framework that integrates frozen backbones, knowledge distillation, and quantization. These techniques reduce communication payload and, consequently, key material consumption. Moving beyond simulation, we benchmark this framework on a real physics-based key distribution testbed involving a chest X-ray classification application. Our results show that key usage can be reduced by $\sim$35$\times$ while maintaining predictive accuracy. This prevents buffer depletion and key expiration, enabling sustainable FL training under physical key generation constraints.
\end{abstract}

\keywords{Federated Learning, Physics-based Key Distribution, Model Compression}

\ccsdesc[500]{Security and privacy~Privacy-preserving protocols}
\ccsdesc[500]{Applied computing~Health informatics}
\ccsdesc[300]{Hardware~Quantum technologies}


\maketitle

\section{Introduction}
Federated Learning (FL) has emerged as a promising paradigm for collaborative model training across institutions that cannot share raw data due to privacy regulations, proprietary concerns, or ethical constraints~\cite{kairouz2021advancesFL}. This is particularly relevant in medical imaging, where individual sites have limited data due to high imaging and annotation costs, yet privacy regulations (e.g., HIPAA, GDPR) prevent pooling datasets across institutions~\cite{guan2024federatedMedicalImaging}. In a standard FL deployment, each participating institution trains a model locally on its own data and transmits only model updates (e.g., gradients or weight deltas) or the updated weights themselves to a central aggregation server. The server aggregates these contributions to produce an improved global model, iterated over multiple communication rounds~\cite{McMahan-17_FL}. While FL avoids raw data exchange, the transmitted model updates or weights are not inherently private: model/gradient inversion attacks~\cite{zhang2020modelInversion,zhu2019deepLeakage} can reconstruct sensitive training samples from shared model/gradients, and membership inference attacks~\cite{shokri2017membership} can determine whether a specific patient's data participated in training. These vulnerabilities are especially acute in medical imaging, where the reconstruction of diagnostic scans directly compromises patient confidentiality.

Secure aggregation protocols address this threat by ensuring that the server learns only the aggregate of client updates, never any individual contribution. A common approach~\cite{bonawitz2017secAggML} uses pairwise additive masking, where each client adds cryptographic masks derived from shared secrets to its model update before transmission. When the server sums all masked updates, the pairwise masks cancel, revealing only the clean aggregate. However, in conventional implementations, the shared secrets are established via classical public-key cryptography (e.g., Diffie-Hellman), which is only computationally secure and vulnerable to attacks by a sufficiently powerful quantum adversary~\cite{shor1994algorithms}. As quantum computing hardware advances, the computational security assumptions underpinning classical secure aggregation become increasingly fragile.

Quantum Key Distribution (QKD) offers a fundamentally different security guarantee~\cite{bennett2014quantum}. Unlike classical key exchange, it encodes key material in quantum states, leveraging the no-cloning theorem and measurement disturbance detection to let two parties establish shared secure keys that are information-theoretically secure, even against an adversary with unlimited computational power. In particular, Continuous-Variable QKD (CV-QKD) with Gaussian modulation encodes key information in the quadratures of coherent states, offering compatibility with standard telecom photonic components~\cite{Laudenbach_2018_CVQKD}.

However, deploying quantum-secure key exchange in FL introduces a fundamental throughput bottleneck. Unlike classical key establishment, which can generate keys at arbitrary rates, information-theoretically secure key generation via quantum channels produces shared secret bits at a finite rate determined by photon detection efficiency, channel loss, and post-processing overhead~\cite{Laudenbach_2018_CVQKD,Xu_2023_edgeqkd}. Furthermore, quantum-derived key material has a limited storage lifetime in practical key management systems, creating a ``use-it-or-lose-it'' constraint~\cite{kaewpuang2022adaptiveresourceallocationquantum}. When Shamir's secret sharing is used for secure aggregation, the key material consumed per round scales linearly with the size of each client's masked model update~\cite{shamir1979share,bonawitz2017secAggML}. For modern medical imaging models (e.g., ResNet-50~\cite{he2015deepresiduallearningimage} with $\sim$\num{25.6e6} parameters), uncompressed weight updates can require hundreds of megabytes of key material per round per client, far exceeding sustainable quantum key generation rates of \qtyrange{1}{10}{\kilo\bit\per\second}~\cite{Brunner2023_QKD}. This mismatch causes key buffer depletion, forcing training to stall. Liu et al.~\cite{liu2026experimentallyvalidatedQKDFL} similarly identify communication cost as the critical bottleneck in quantum-secure FL, achieving \qty{75}{\percent} reduction through model compression, yet still do not address sustained multi-round key consumption for large architectures. This fundamental tension limits the feasibility of quantum-secured FL for realistic medical imaging workloads.

Recent work has begun exploring quantum-secured FL.\@ Catak et al.~\cite{catak2026quantumkeydistributionsecured} demonstrate quantum-secured FL for 6G channel estimation with BB84-style key distribution. Liu et al.~\cite{liu2026experimentallyvalidatedQKDFL} provide the first experimental validation on a physical four-client quantum network, achieving \qty{75}{\percent} communication cost reduction via model compression. Tanbhir and Shahriyar~\cite{Tanbhir_2025_qkd_medical_imaging} integrate quantum key exchange with FL for dementia classification on brain MRI.\@ Broader surveys have also begun to appear~\cite{nguyen2025quantumfederatedlearningcomprehensive, Mathur_2026_qflsurvey}. However, these works either do not address sustained key consumption over multiple FL rounds, do not evaluate on realistic workloads (such as those that are typical in medical imaging \cite{khan2024accuracy,han2025federated}), or remain simulation-based. To our knowledge, no prior work systematically addresses the tension between quantum key generation constraints and FL payload size through compression on a real quantum key testbed.

In this work, we bridge this gap by developing an FL framework that systematically reduces key material consumption to match quantum key throughput constraints. We combine three complementary techniques: (i) frozen pretrained backbones that eliminate the need to transmit large feature-extractor weights, reducing the number of trainable (and thus transmittable) parameters; (ii) knowledge distillation to compress models into low-dimensional student representations; and (iii) quantization to reduce the bit-width of transmitted parameters. We deploy this pipeline on a real CV quantum key testbed and evaluate it on a public dataset related to chest X-ray classification, demonstrating that key usage can be reduced sufficiently to prevent buffer depletion and key expiration while maintaining accuracy.

Overall, our main contributions can be summarized as follows:
\begin{itemize}[leftmargin=*]
    \item We propose a communication-efficient FL framework combining frozen backbones, knowledge distillation, and quantization that reduces per-round key material consumption to levels compatible with practical quantum key generation rates.
    \item We deploy and evaluate the framework on a real CV quantum key testbed for chest X-ray classification, providing the first experimental demonstration that quantum-secured FL can sustain multi-round training without key buffer depletion.
    \item We quantify the trade-off between communication compression, accuracy, and key sustainability.
\end{itemize}

\section{Proposed Framework}\label{sec:framework}

\begin{figure}[t]
    \centering
    \includegraphics[width=.5\textwidth]{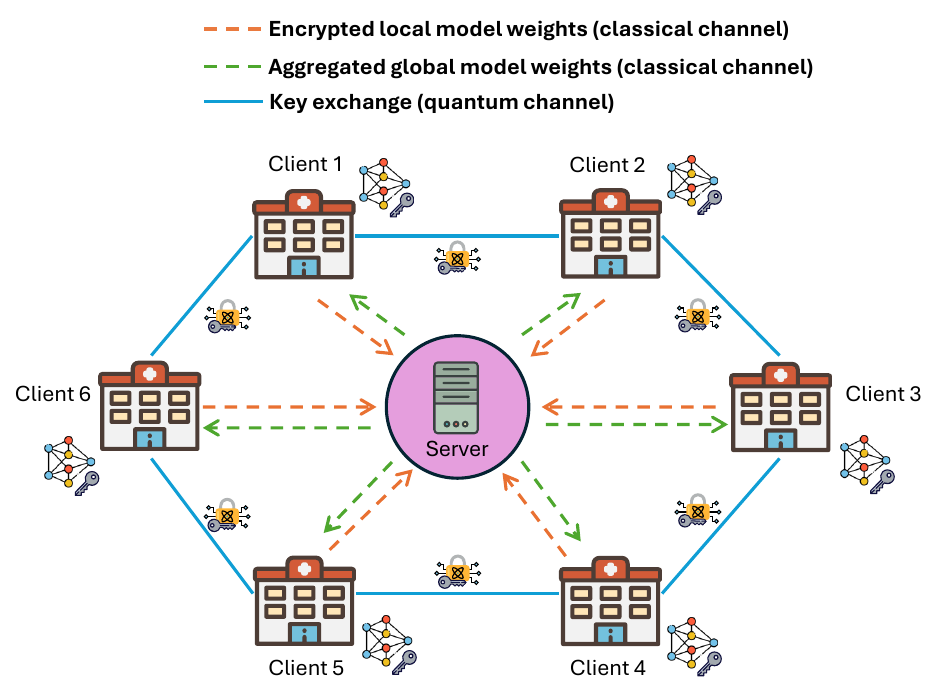}
    \caption{Overview of the proposed FL framework. Each client trains a classification head locally (backbone frozen), quantizes the updated parameters, and applies additive masks derived from quantum key material shared with ring neighbors. The server sums all masked updates (pairwise masks cancel exactly) and updates the global model via FedAvg.}\label{fig:fl_network}
    \Description{Overview of the proposed FL framework.}
\end{figure}

The proposed framework integrates quantum-secured additive masking with model compression to enable sustainable FL under
physical key generation constraints. \Cref{fig:fl_network} illustrates the end-to-end flow. Each client trains a classification
head locally on its data partition while keeping the backbone frozen. 
The trained head parameters are quantized to fixed-width
integers. Each client then fetches key material (shared with its two ring neighbors) and applies additive or subtractive masks
to the quantized parameters. The masked vectors are transmitted to the server, which sums them; the pairwise masks cancel exactly,
yielding the clean aggregate. The server dequantizes the aggregate and updates the global model via Federated Averaging (FedAvg)~\cite{McMahan-17_FL}.

\subsection{Quantum-Secured Ring Additive Masking}\label{sec:ring_masking}
\noindent\textbf{Threat model.}
The primary adversary is an honest-but-curious aggregation server with unbounded computational power.
The server follows the protocol faithfully but attempts to infer individual client contributions from the masked updates it receives.
We assume the pairwise keys derived via QKD are information-theoretically secure,
that each key masks exactly one update and is never reused,
and that classical components (key storage, transport, software) are uncompromised;
implementation-level side channels are out of scope (\Cref{sec:discussion}).
Security is assessed at three levels:
\begin{enumerate}[leftmargin=*,itemsep=1pt,topsep=2pt]
\item \textbf{Key security:}
the pairwise keys are information-theoretically secure, a guarantee inherited from the physical layer~\cite{diamanti2016practical}.
\item \textbf{Message security:}
given key security, each masked update is statistically independent of the underlying contribution, the keys act as one-time pads~\cite{shannon1949secrecy}.
The server alone therefore learns nothing about any individual contribution beyond the aggregate.
This guarantee also holds if the server colludes with any single client,
even a ring neighbor of the target, since reconstruction requires both neighboring keys.
\item \textbf{System-level privacy:}
client compromise is outside the primary model, but its consequences can be stated precisely.
A coalition comprising the server and both ring neighbors of a client can reconstruct that client's contribution.
This is the price of reducing quantum links from $O(n^2)$ to $O(n)$;
fully connected pairwise masking tolerates collusion of up to $n-1$ clients~\cite{bonawitz2017secAggML}.
\end{enumerate}

We arrange $n$ clients indexed by $[n] = \{1, 2, \ldots, n\}$ in a ring topology, where each client $i$ is connected to its neighbors $i-1$ and $i+1$ via dedicated links. Each link generates shared secure key material between the two endpoints with information-theoretic security. Key material is accumulated into per-link buffers (\Cref{sec:deployment}) and drawn upon as needed. Prior to each FL round, client $i$ fetches two keys from its local key buffer: $K_{i-1,i}$, shared with client $i-1$, and $K_{i,i+1}$, shared with client $i+1$.
To preserve information-theoretic security in the masking layer, these raw key bytes are used directly as additive masks.
The keys are reshaped to match the dimensionality of the quantized model update $\theta_i$.

The masking scheme works as follows. Each client $i$ adds the key shared with its right neighbor ($K_{i,i+1}$) and subtracts the key shared with its left neighbor ($K_{i-1,i}$). Because each key is shared by exactly two adjacent clients, it is added once and subtracted once across the ring, causing all masks to cancel exactly when the server sums all masked updates. Formally, the client computes its masked update:
\begin{equation}
    \hat{\theta}_i = \theta_i + K_{i,i+1} - K_{i-1,i} \pmod{2^{32}},
\end{equation}
and the server computes the global aggregate:
\begin{equation}
    \sum_{i=1}^{n} \hat{\theta}_i = \sum_{i=1}^{n} \theta_i + \sum_{i=1}^{n} \left( K_{i,i+1} - K_{i-1,i} \right) = \sum_{i=1}^{n} \theta_i \pmod{2^{32}},
\end{equation}
where all pairwise masks vanish because each key appears once with a positive sign and once with a negative sign, leaving the server with only the clean aggregate $\sum_{i=1}^{n} \theta_i$ without access to any individual contribution.

Because the masks are drawn directly from the quantum key material without pseudorandom expansion, the level-2 guarantee holds regardless of the adversary's computational power: no computational assumption (Diffie–Hellman hardness or pseudorandomness) remains in the masking layer, unlike in classical secure aggregation.

The ring topology reduces the number of required links from $O(n^2)$ to $O(n)$, which is critical given the cost of QKD hardware. The hardware implementation, which further halves the device count by sharing transmitters and receivers across links through time-division multiplexing, is described in \Cref{sec:deployment}. Because the masks are not expanded, the key consumption per client per round is exactly $2 \times P$, where $P$ is the compressed payload size. This makes the compression pipeline described in the next section essential for sustaining training without depleting the key buffers.

\subsection{Compression Techniques}\label{sec:compression}
We apply a three-stage compression pipeline to reduce the payload size transmitted during FL rounds.

\subsubsection{Frozen Pretrained Backbone}
All backbone layers are frozen, and only the classification head is trained and transmitted. The model architecture considered in our experiments, i.e., MobileNetV3-Small~\cite{howard2019searchingmobilenetv3} (width multiplier $0.75$) pretrained on ImageNet\cite{imagenet2009}, comprises $\sim$\num{803e3} parameters; whereas the classification head comprises $\sim$\num{229e3} parameters, yielding a $\sim$3.5$\times$ reduction in the number of parameters that are trained and transmitted.

\subsubsection{Knowledge Distillation}
To improve the quality of the frozen features, we employ knowledge distillation \cite{bilal2025enhancing} as a domain-transfer pretraining strategy. A compact student model ($\sim$\num{337e3} total parameters, $\sim$\num{90e3} for the classification head) is distilled from a MobileNetV3-Small (width multiplier $1$) teacher fine-tuned on the CheXpert dataset~\cite{irvin2019chexpertlargechestradiograph}\footnote{CheXpert,
available from Stanford AIMI, \url{https://doi.org/10.71718/y7pj-4v93}}, producing domain-specific thoracic features that outperform ImageNet features for chest X-ray classification. The student is then frozen, and also in this case only its classification head is transmitted, reducing the payload by a further $\sim$2.5$\times$ relative to the frozen teacher head.

\subsubsection{Quantization}
Transmitted model updates are quantized to fixed-width integers using a symmetric clipping range of $[-8.0, +8.0]$. We evaluate three precision levels: 32-bit, 16-bit, and 8-bit. Quantization is applied only to the parameters transmitted to the aggregation server; local training remains in float32 to preserve gradient quality.

\section{Experimental Setup}\label{sec:setup}

\subsection{Dataset and Partitioning}
We evaluate on the NIH ChestX-ray14 dataset~\cite{wang2017NIHchestxray}, comprising 112,120 frontal-view chest X-rays from 30,805 unique patients with 15 multi-label thoracic disease classes. Images are resized to $256{\times}256$. Then, the following augmentations are applied: random crop to $224{\times}224$, horizontal flip ($p{=}0.5$), $\pm10^\circ$ rotation, affine (translate $\pm5\%$, scale $0.95$--$1.05$), brightness/contrast jitter $\pm0.2$.  Next, the images are normalized using ImageNet channel-wise statistics ($\mu = (0.485, 0.456, 0.406)$, $\sigma = (0.229, 0.224, 0.225)$)\cite{imagenet2009}. We use the official train/test split (86,524 train / 25,596 test images), with patient-level, verified-disjoint partitioning: each client additionally splits its patients 85\%/15\% into train/validation. IID clients each receive exactly 4,668 patients (${\sim}14{,}400$ images); Dirichlet($\alpha{=}0.1$) label-skew clients range from ${\sim}7{,}100$ to ${\sim}23{,}000$ images (draws rejected below 800 training images per client). Training is monitored via client-side validation; the test set is evaluated only after the final round.
For knowledge distillation pretraining, as seen earlier, we use the CheXpert dataset (224,316 chest radiographs with 14 pathology classes and uncertainty annotations). Uncertain labels ($-1$) are mapped to positive following the ``ones'' policy~\cite{irvin2019chexpertlargechestradiograph}.

\subsection{Model Architecture}
We use MobileNetV3-Small~\cite{howard2019searchingmobilenetv3} as the backbone architecture, chosen for CPU-only deployment at client sites. Two model configurations are evaluated:
\begin{itemize}[leftmargin=*]
    \item \textbf{Teacher.} MobileNetV3-Small (width multiplier $0.75$) pretrained on ImageNet, used either fully trainable (uncompressed baseline) or as a frozen feature extractor. The full model comprises $\sim$\num{803e3} parameters; the classifier head comprises $\sim$\num{229e3} parameters. The size of the classification head was halved w.r.t.\ the original MobileNetV3, since the target task has only 15 classes instead of the 1000 used for the ImageNet pretraining.

    \item \textbf{Student.} A compact network derived from the MobileNetV3 architecture by reducing channel width and network depth ($\sim$\num{337e3} parameters total, \num{90e3} parameters for the classification head), distilled from a MobileNetV3-Small (width multiplier $1$) teacher fine-tuned on CheXpert. The distillation loss combines temperature-scaled per-class Bernoulli KL divergence on sigmoid soft targets with masked binary cross-entropy:
    \begin{equation}
    \mathcal{L} = \alpha\,\mathrm{KL}_{\mathrm{Bern}}\!\left(
    \sigma\!\left(\tfrac{f_S}{T}\right) \,\Big\|\,
    \sigma\!\left(\tfrac{f_T}{T}\right)\right)
    + (1-\alpha)\,\mathrm{BCE}(y, f_S),
    \end{equation}
    where $\mathrm{BCE}$ is Binary Cross Entropy, $f_S$ and $f_T$ are student and teacher logits, respectively, $T$ is the temperature, $\alpha$ controls the distillation weight, and $\sigma$ is the sigmoid function. The student was distilled with learning rate ${\eta}=10^{-4}$, temperature $T=8$ and $\alpha=0.7$, achieving mean AUC $= 0.761$ on the CheXpert validation set.
\end{itemize}

\subsection{Federated Learning Configuration}
We implement the FL framework using the Flower framework~\cite{flower2020}. All experiments are conducted on a cluster of six physical machines, each representing one FL client. Each node is equipped with dual Intel Xeon E5-2683 v4 processors (64 logical CPUs total) and \qty{125}{\gibi\byte} RAM, with training parallelized across 12 logical cores per client. All training is performed on CPU only; no GPUs are used. This CPU-only configuration reflects realistic deployment constraints at clinical sites, where hospitals typically utilize existing enterprise server infrastructure rather than provisioning specialized AI accelerators. Six clients are arranged in a ring topology implemented with three transmitters and three receivers (\Cref{sec:ring_masking}). We run FedAvg for 50 communication rounds with 1 local epoch per round, batch size 16, starting learning rate $10^{-2}$ with a cosine annealing learning rate scheduler, and weight decay $10^{-4}$. All experiments with masking use frozen backbones with only the classification head trainable. Under this configuration, each FL round completes in approximately \qty{4}{\minute}.

\subsection{Quantum Key Distribution Network}\label{sec:deployment}

The testbed comprises three CV-QKD transmitters and three receivers co-located in a laboratory. The used prototype devices are field-trial tested as described in~\cite{Brunner2023_QKD}. The transmitters operate at fixed carrier frequencies in the C-band (\qty{193.4}{\tera\hertz}, \qty{193.7}{\tera\hertz}, and \qty{193.8}{\tera\hertz}). Each transmitter output is split via a 1:2 coupler to its two neighboring receivers; each receiver input is a combination of the two neighboring transmitters via a 2:1 coupler. A receiver selects the desired signal by tuning its local oscillator wavelength. The six logical links form a ring topology, as illustrated in~\cite{Mendez_2026_QKD}.

Because each receiver can only detect one signal at a time, only three of the six links are active simultaneously. The odd and even link sets alternate approximately every hour, coordinated by a switching controller~\cite{Mendez_2026_QKD}. Despite this \qty{50}{\percent} duty cycle, key material is continuously available through per-link key buffers: each buffer holds up to 200{,}000 keys of \qty{512}{\byte} each ($\sim$\qty{100}{\mega\byte}).

Achievable key generation rates for CV-QKD networks of this type range from \qtyrange{1}{10}{\kilo\bit\per\second} for channel losses of \qtyrange{4}{16}{\decibel}~\cite{Brunner2023_QKD}. The present testbed operates at the upper end of this range ($\sim$\qty{10}{\kilo\bit\per\second} per active link). While the current deployment is a co-located laboratory setup with prototype devices, a demonstration over a metropolitan fiber network is planned as future work.

\subsection{Evaluation Metrics}
We report the macro-averaged one-versus-rest AUROC score.

\section{Experimental Results}\label{sec:results}

\subsection{Classification Performance}\label{sec:cls_results}
\Cref{tab:results} reports AUROC (macro-averaged) on the NIH ChestX-ray14 test set after 50 FL rounds across all configurations. The uncompressed teacher baseline (fully trainable, no masking) achieves an AUROC of 0.762 under IID partitioning. Freezing the backbone and transmitting only the classification head yields a modest accuracy decrease (0.745 IID, 0.736 non-IID), consistent with the loss of fine-tuning in the feature extractor. As an implementation sanity check, we repeated the frozen-teacher configuration with masking disabled: the per-round aggregates and the final AUROC (0.745) are identical to the masked run, confirming that the pairwise masks cancel exactly and that the masking layer does not alter the training trajectory. The CheXpert-distilled student recovers much of this gap (0.748 IID), demonstrating that domain-specific distillation compensates for the reduced model capacity. Quantization to 16-bit and 8-bit incurs a modest additional degradation (up to $\sim$0.02 AUROC from 32-bit to 8-bit), confirming that fixed-width integer representation is viable for FL weight transmission. Non-IID partitioning reduces AUROC by $\sim$0.01--0.02 across all configurations, as expected.
\begin{table}[t]
\centering
\small
\caption{Macro-averaged AUROC on the NIH ChestX-ray14 test set after 50 FL rounds.
The teacher is ImageNet-pretrained, and the student is CheXpert-distilled.}\label{tab:results}
\begin{tabularx}{\columnwidth}{>{\raggedright\arraybackslash}X >{\centering\arraybackslash}X >{\centering\arraybackslash}X >{\centering\arraybackslash}X >{\centering\arraybackslash}X >{\centering\arraybackslash}X}
\toprule
Model & Frozen & Quant. & Partition & Payload & AUROC \\
\midrule
Teacher (No masking) & No  & 32-bit & IID    & \qty{3.2}{\mega\byte}  & 0.762 \\
\addlinespace
Teacher (No masking) & Yes & 32-bit & IID    & \qty{916}{\kilo\byte} & 0.745 \\
Teacher & Yes & 32-bit & IID    & \qty{916}{\kilo\byte} & 0.745 \\
Teacher & Yes & 32-bit & non-IID & \qty{916}{\kilo\byte} & 0.736 \\
\addlinespace
Student & Yes & 32-bit & IID    & \qty{360}{\kilo\byte} & 0.748 \\
Student & Yes & 16-bit & IID    & \qty{180}{\kilo\byte} & 0.738 \\
Student & Yes & 8-bit  & IID    & \qty{90}{\kilo\byte}  & 0.732 \\
\addlinespace
Student & Yes & 32-bit & non-IID & \qty{360}{\kilo\byte} & 0.731 \\
Student & Yes & 16-bit & non-IID & \qty{180}{\kilo\byte} & 0.719 \\
Student & Yes & 8-bit  & non-IID & \qty{90}{\kilo\byte}  & 0.712 \\
\bottomrule
\end{tabularx}
\end{table}

\subsection{Key Consumption and Sustainability}\label{sec:sustainability}
Each client consumes $2P$ key material per round, one payload-sized key per ring neighbor; since each link's key is drawn by both endpoints, this corresponds to $P$ per link and $nP$ system-wide. Payload sizes $P$ for all evaluated configurations are given in \Cref{tab:results}; relative to the uncompressed teacher (\qty{3.2}{\mega\byte}), the 8-bit student head cuts per-round key
consumption by ${\sim}35\times$. Given the measured active-link rate of ${\sim}$\qty{10}{\kilo\bit\per\second} and the \qty{50}{\percent} duty cycle, each link generates ${\sim}$\qty{150}{\kilo\byte} per \qty{4}{\minute} round; \Cref{tab:sustainability} compares this budget against consumption. The 8-bit student is indefinitely sustainable; the remaining masked configurations drain the pre-charged ${\sim}$\qty{100}{\mega\byte} buffers but complete all 50 rounds (worst-case deficit ${\sim}$\qty{38}{\mega\byte}), whereas the uncompressed baseline would exhaust a full buffer after ${\sim}$33 rounds. Sustainability scales linearly with key rate and round duration ($P \le \bar{R}\,T$ per link): at $\bar{R}=\qty{5}{\kilo\bit\per\second}$, the 8-bit student sustains rounds longer than \qty{144}{\second}, but a \qty{60}{\second} GPU-paced round would require \qty{12}{\kilo\bit\per\second} on average (\qty{24}{\kilo\bit\per\second} at \qty{30}{\second}), beyond the \qty{10}{\kilo\bit\per\second} active-link rate of the present testbed. Faster client hardware would thus require more local computation per round, sparser aggregation, or further compression to restore the margin.

\begin{table}[t]
\centering
\small
\setlength{\tabcolsep}{2.5pt}
\caption{Per-link key budget per round at average key rate $\bar{R}\approx\qty{5}{\kilo\bit\per\second}$ and round duration $T\approx\qty{4}{\minute}$: consumption, net balance per round, minimum sustainable round duration $T_{\min}=P/\bar{R}$ (in minutes), and rounds sustained by a full ${\sim}$\qty{100}{\mega\byte} link buffer. For the non-frozen teacher, it shows the key consumption it would incur if masked; consumption is identical under IID and non-IID partitioning.}\label{tab:sustainability}
\begin{tabularx}{\columnwidth}{>{\raggedright\arraybackslash}X >{\centering\arraybackslash}X >{\centering\arraybackslash}X >{\centering\arraybackslash}X >{\centering\arraybackslash}X >{\centering\arraybackslash}X >{\centering\arraybackslash}X}
\toprule
Model & Frozen & Quant. & Consumed & Net & $T_{\min}$ & Rounds \\
\midrule
Teacher & No & 32-bit & \qty{3.2}{\mega\byte} & $-$\qty{3.05}{\mega\byte} & $\sim$85 & $\sim$33 \\
Teacher & Yes & 32-bit & \qty{916}{\kilo\byte} & $-$\qty{766}{\kilo\byte} & $\sim$24 & $\sim$130 \\
Student & Yes & 32-bit & \qty{360}{\kilo\byte} & $-$\qty{210}{\kilo\byte} & $\sim$9.6 & $\sim$480 \\
Student & Yes & 16-bit & \qty{180}{\kilo\byte} & $-$\qty{30}{\kilo\byte} & $\sim$4.8 & $\sim$3{,}300 \\
Student & Yes & 8-bit & \qty{90}{\kilo\byte} & $+$\qty{60}{\kilo\byte} & 2.4 & Indefinite \\
\bottomrule
\end{tabularx}
\end{table}

\section{Discussion}
\label{sec:discussion}

\textbf{Key findings.}
The results show that quantum-secured FL is practically feasible on real hardware when paired with aggressive compression. The uncompressed baseline would consume \qty{3.2}{\mega\byte} per round per link ($T_{\min}\approx\qty{85}{\minute}$, \Cref{tab:sustainability}), exhausting a full buffer within ${\sim}33$ rounds, whereas the 8-bit student needs only \qty{90}{\kilo\byte} per link ($T_{\min}\approx\qty{2.4}{\minute}$) and is indefinitely sustainable.

\textbf{Accuracy trade-offs.}
The choice of pretraining strategy proves critical for maintaining accuracy under heavy compression. The ImageNet-pretrained backbone exhibits a substantial domain shift when applied to chest radiographs, resulting in a lower AUC compared to the domain-adapted model. By using knowledge distillation from a teacher fine-tuned on the CheXpert dataset, the student model acquires thoracic-specific features that transfer seamlessly to the NIH dataset task. Once this high-quality frozen backbone is established, the trainable classification head requires very few parameters. Consequently, quantizing the transmitted head parameters from 32-bit to 8-bit integers introduces only marginal accuracy degradation. This confirms that the bulk of the predictive capability resides in the stable, frozen feature extractors rather than the rapidly updated classification head, justifying the aggressive payload reduction.

\textbf{Non-IID robustness.}
Real-world medical FL deployments face naturally non-IID data due to varying disease prevalence across hospitals. Our framework maintains stable convergence under aggressive label skew ($\alpha=0.1$) with only modest accuracy degradation relative to IID. Because the frozen, domain-pretrained backbone confines local updates to the linear head, the restricted gradient space reduces divergence risk, suggesting the compression pipeline does not amplify non-IID optimization challenges.

\textbf{Security considerations.} 
The levels-1 and -2 security guarantees are information-theoretic in principle~\cite{shannon1949secrecy, diamanti2016practical},
but practical deployments introduce implementation-level vulnerabilities that fall outside this model,
including side-channel attacks on quantum key exchange hardware and classical attack surfaces in the network transport and software stack.
The framework's contribution is therefore best understood as a strict improvement over classical secure aggregation with respect to key establishment and mask generation: it removes the computational assumptions (Diffie-Hellman hardness and pseudorandomness) while retaining the same trust assumption about the aggregation server. Collusion resistance (level-3) is weaker
than in fully connected pairwise masking, as discussed above. A formal security analysis incorporating implementation-level threats is left to future work.

\textbf{Limitations.}
The ring additive masking scheme provides information-theoretic security with minimal quantum key exchange infrastructure, but introduces two structural constraints. First, it requires all clients to participate in every round; if a single client drops out or becomes unresponsive, the pairwise masks do not cancel at the server, and the aggregate cannot be recovered. Dropout-resilient mechanisms, such as having the server reveal the dropped client's mask or employing threshold secret sharing variants, could address this in future work. Second, the ring topology reduces collusion resistance relative to fully connected pairwise masking: a coalition of the server and both ring neighbors of a client can unmask its contribution (threat-model level 3). This is a direct consequence of minimizing the number of quantum key links from $O(n^2)$ to $O(n)$, and we consider it an acceptable trade-off given that the primary threat model is an honest-but-curious server. Strengthening the scheme against collusion, e.g., through multi-hop masking or hybrid topologies, is left to future work. Finally, the framework is currently evaluated on a single dataset (NIH ChestX-ray14) and a single architecture family (MobileNetV3). While the approach generalizes conceptually to other medical imaging modalities and backbone choices, empirical validation across diverse tasks remains necessary.

\section{Conclusions}
\label{sec:conclusions}
In this work, we addressed the fundamental throughput bottleneck that arises when integrating QKD with FL. While it provides information-theoretically secure key exchange, its finite generation rates and perishable key buffers are fundamentally incompatible with the massive payload sizes of modern medical imaging models. We proposed a framework that bridges this gap by combining frozen pretrained backbones, knowledge distillation for domain adaptation, and fixed-width quantization. Deployed on a real CV quantum key testbed with six clients arranged in a ring topology, the framework reduced per-round key material consumption by up to $\sim$35$\times$. This reduction enabled sustained, 50-round federated training on chest X-ray classification without key buffer depletion, all while preserving accuracy. Future work will focus on integrating dropout resilience into the masking protocol and extending the evaluation to additional medical imaging modalities and backbone architectures, or alternative FL aggregation functions, such as those based on ensembles \cite{cansiz2026evolutionary}.

\bibliographystyle{ACM-Reference-Format}
\bibliography{bibliography}

@string{Computer = "{IEEE} Computer" }

@string{Springer = "Springer-Verlag" }

@article{shannon1949secrecy,
	title        = {{Communication theory of secrecy systems}},
	author       = {Shannon, C. E.},
	year         = 1949,
	journal      = {The Bell System Technical Journal},
	volume       = 28,
	number       = 4,
	pages        = {656--715},
	doi          = {10.1002/j.1538-7305.1949.tb00928.x},
}

@article{diamanti2016practical,
	title        = {{Practical challenges in quantum key distribution}},
	author       = {Diamanti, Eleni and Lo, Hoi-Kwong and Qi, Bing and Yuan, Zhiliang},
	year         = 2016,
	month        = {Nov},
	journal      = {npj Quantum Information},
	publisher    = {Springer Science and Business Media LLC},
	volume       = 2,
	number       = 1,
    numpages     = 12,
	doi          = {10.1038/npjqi.2016.25},
	issn         = {2056-6387},
	url          = {http://dx.doi.org/10.1038/npjqi.2016.25}
}

@article{bennett2014quantum,
	title        = {{Quantum cryptography: Public key distribution and coin tossing}},
	author       = {Bennett, Charles H and Brassard, Gilles},
	year         = 2014,
	journal      = {Theoretical computer science},
	publisher    = {Elsevier},
	volume       = 560,
	pages        = {7--11}
}

@inproceedings{shor1994algorithms,
	title        = {{Algorithms for quantum computation: discrete logarithms and factoring}},
	author       = {Shor, Peter W},
	year         = 1994,
	booktitle    = {IEEE Annual Symposium on Foundations of Computer Science},
	pages        = {124--134},
    publisher    = {IEEE},
    address      = {New York, NY, USA},
}

@article{Laudenbach_2018_CVQKD,
	title        = {{Continuous‐Variable Quantum Key Distribution with Gaussian Modulation—The Theory of Practical Implementations}},
	author       = {Laudenbach, Fabian and Pacher, Christoph and Fung, Chi‐Hang Fred and Poppe, Andreas and Peev, Momtchil and Schrenk, Bernhard and Hentschel, Michael and Walther, Philip and Hübel, Hannes},
	year         = 2018,
	journal      = {Advanced Quantum Technologies},
	publisher    = {Wiley},
    address      = {Hoboken, NJ, USA},
	volume       = 1,
	number       = 1,
    pages        = {1800011},
	doi          = {10.1002/qute.201800011},
	issn         = {2511-9044},
	url          = {http://dx.doi.org/10.1002/qute.201800011}
}

@article{Mendez_2026_QKD,
	title        = {{Switching Coordinator: An SDN Application for Flexible QKD Networks}},
	author       = {Méndez, Rubén B. and Brunner, Hans H. and Brito, Juan P. and Taramit, Hamid and Fung, Chi-Hang Fred and Pastor, Antonio and Cantó, Rafael and Folgueira, Jesús and López, Diego R. and Peev, Momtchil and Martin, Vicente},
	year         = 2026,
	month        = {Feb},
	journal      = {Entropy},
	publisher    = {MDPI AG},
	volume       = 28,
	number       = 2,
	pages        = 219,
	doi          = {10.3390/e28020219},
	issn         = {1099-4300},
	url          = {http://dx.doi.org/10.3390/e28020219}
}

@article{Brunner2023_QKD,
	title        = {{Demonstration of a switched {CV-QKD} network}},
	author       = {Brunner, Hans H. and Fung, Chi-Hang Fred and Peev, Momtchil and Méndez, Rubén B. and Ortiz, Laura and Brito, Juan P. and Martín, Vicente and Rivas-Moscoso, José M. and Jiménez, Felipe and Pastor, Antonio A. and López, Diego R.},
	year         = 2023,
	month        = sep,
	journal      = {EPJ Quantum Technology},
	volume       = 10,
	number       = 1,
	pages        = 38,
	doi          = {10.1140/epjqt/s40507-023-00194-x},
	issn         = {2196-0763},
	url          = {https://doi.org/10.1140/epjqt/s40507-023-00194-x}
}

@inproceedings{McMahan-17_FL,
	title        = {{Communication-Efficient Learning of Deep Networks from Decentralized Data}},
	author       = {B. McMahan and E. Moore and D. Ramage and S. Hampson and B. A. y Arcas},
	year         = 2017,
	booktitle    = {International Conference on Artificial Intelligence and Statistics (AISTATS'17)},
	publisher    = {PMLR},
    address      = {Fort Lauderdale, FL, USA},
	pages        = {1273--1282}
}

@article{kairouz2021advancesFL,
	title        = {{Advances and open problems in federated learning}},
	author       = {Kairouz, Peter and McMahan, H Brendan},
	year         = 2021,
	journal      = {Foundations and trends in machine learning},
	publisher    = {Emerald Publishing Limited},
	volume       = 14,
	number       = {1-2},
	pages        = {1--210}
}

@article{guan2024federatedMedicalImaging,
	title        = {{Federated learning for medical image analysis: A survey}},
	author       = {Guan, Hao and Yap, Pew-Thian and Bozoki, Andrea and Liu, Mingxia},
	year         = 2024,
	journal      = {Pattern recognition},
	publisher    = {Elsevier},
	volume       = 151,
	pages        = 110424
}

@misc{flower2020,
	title        = {{Flower: {A} Friendly Federated Learning Research Framework}},
	author       = {Daniel J. Beutel and Taner Topal and Akhil Mathur and Xinchi Qiu and Titouan Parcollet and Nicholas D. Lane},
	year         = 2020,
	volume       = {abs/2007.14390},
	url          = {https://arxiv.org/abs/2007.14390},
	eprinttype   = {arXiv},
	eprint       = {2007.14390},
	bibsource    = {dblp computer science bibliography, https://dblp.org}
}

@inproceedings{shokri2017membership,
	title        = {{Membership inference attacks against machine learning models}},
	author       = {Shokri, Reza and Stronati, Marco and Song, Congzheng and Shmatikov, Vitaly},
	year         = 2017,
	booktitle    = {IEEE Symposium on Security and Privacy (SP)},
	pages        = {3--18},
	publisher    = {IEEE},
    address      = {New York, NY, USA},
}

@inproceedings{zhang2020modelInversion,
	title        = {{The secret revealer: Generative model-inversion attacks against deep neural networks}},
	author       = {Zhang, Yuheng and Jia, Ruoxi and Pei, Hengzhi and Wang, Wenxiao and Li, Bo and Song, Dawn},
	year         = 2020,
	booktitle    = {IEEE/CVF Conference on Computer Vision and Pattern Recognition},
    publisher    = {IEEE},
    address      = {New York, NY, USA},
	pages        = {253--261}
}

@article{zhu2019deepLeakage,
	title        = {{Deep leakage from gradients}},
	author       = {Zhu, Ligeng and Liu, Zhijian and Han, Song},
	year         = 2019,
	journal      = {Advances in neural information processing systems},
	volume       = 32,
    numpages     = {11},
    publisher    = {Curran Associates Inc.},
    address      = {Red Hook, NY, USA},
}

@article{shamir1979share,
	title        = {{How to share a secret}},
	author       = {Shamir, Adi},
	year         = 1979,
	journal      = {Communications of the ACM},
	publisher    = {ACM},
    address      = {New York, NY, USA},
	volume       = 22,
	number       = 11,
	pages        = {612--613}
}

@inproceedings{bonawitz2017secAggML,
	title        = {{Practical secure aggregation for privacy-preserving machine learning}},
	author       = {Bonawitz, Keith and Ivanov, Vladimir and Kreuter, Ben and Marcedone, Antonio and McMahan, H Brendan and Patel, Sarvar and Ramage, Daniel and Segal, Aaron and Seth, Karn},
	year         = 2017,
	booktitle    = {ACM SIGSAC Conference on Computer and Communications Security},
    publisher    = {ACM},
    address      = {New York, NY, USA},
	pages        = {1175--1191}
}

@misc{nguyen2025quantumfederatedlearningcomprehensive,
	title        = {{Quantum Federated Learning: A Comprehensive Survey}},
	author       = {Dinh C. Nguyen and Md Raihan Uddin and Shaba Shaon and Ratun Rahman and Octavia Dobre and Dusit Niyato},
	year         = 2025,
	url          = {https://arxiv.org/abs/2508.15998},
	eprint       = {2508.15998},
	archiveprefix = {arXiv},
	primaryclass = {cs.LG}
}

@article{Mathur_2026_qflsurvey,
	title        = {{When Federated Learning Meets Quantum Computing: Survey and Research Opportunities}},
	author       = {Mathur, Aakar and Gupta, Ashish and Das, Sajal K.},
	year         = 2026,
	journal      = {IEEE Communications Surveys \& Tutorials},
	publisher    = {Institute of Electrical and Electronics Engineers (IEEE)},
	volume       = 28,
	pages        = {1351–1380},
	doi          = {10.1109/comst.2025.3634143},
	issn         = {2373-745X},
	url          = {http://dx.doi.org/10.1109/COMST.2025.3634143}
}

@misc{catak2026quantumkeydistributionsecured,
	title        = {{Quantum Key Distribution Secured Federated Learning for Channel Estimation and Radar Spectrum Sensing in 6G Networks}},
	author       = {Ferhat Ozgur Catak and Murat Kuzlu and Jungwon Seo and Umit Cali},
	year         = 2026,
	url          = {https://arxiv.org/abs/2603.15649},
	eprint       = {2603.15649},
	archiveprefix = {arXiv},
	primaryclass = {cs.CR}
}

@article{liu2026experimentallyvalidatedQKDFL,
	title        = {{Experimentally Validated Quantum-Secure Federated Learning over a Multi-user Quantum Network}},
	author       = {Zhi-Ping Liu  and Xiao-Yu Cao  and Hao-Wen Liu  and Xiao-Ran Sun  and Yu Bao  and Jian-Yu Shen  and Yu-Shuo Lu  and Hua-Lei Yin  and Zeng-Bing Chen},
	year         = 2026,
	journal      = {Research},
	volume       = 9,
	pages        = 1299,
	doi          = {10.34133/research.1299},
	url          = {https://spj.science.org/doi/abs/10.34133/research.1299},
	eprint       = {https://spj.science.org/doi/pdf/10.34133/research.1299}
}

@inproceedings{Tanbhir_2025_qkd_medical_imaging,
	title        = {{Quantum-Inspired Privacy-Preserving Federated Learning Framework for Secure Dementia Classification}},
	author       = {Tanbhir, Gazi and Shahriyar, Md Farhan},
	year         = 2025,
	month        = {Feb},
	booktitle    = {IEEE International Conference on Electrical, Computer and Communication Engineering (ECCE)},
	publisher    = {IEEE},
    address      = {New York, NY, USA},
	pages        = {1–6},
	doi          = {10.1109/ecce64574.2025.11013884},
	url          = {http://dx.doi.org/10.1109/ECCE64574.2025.11013884}
}

@misc{kaewpuang2022adaptiveresourceallocationquantum,
	title        = {{Adaptive Resource Allocation in Quantum Key Distribution (QKD) for Federated Learning}},
	author       = {Rakpong Kaewpuang and Minrui Xu and Dusit Niyato and Han Yu and Zehui Xiong and Xuemin Sherman Shen},
	year         = 2022,
	url          = {https://arxiv.org/abs/2208.11270},
	eprint       = {2208.11270},
	archiveprefix = {arXiv},
	primaryclass = {cs.CR}
}

@article{Xu_2023_edgeqkd,
	title        = {{Privacy-Preserving Intelligent Resource Allocation for Federated Edge Learning in Quantum Internet}},
	author       = {Xu, Minrui and Niyato, Dusit and Yang, Zhaohui and Xiong, Zehui and Kang, Jiawen and Kim, Dong In and Shen, Xuemin},
	year         = 2023,
	month        = {Jan},
	journal      = {IEEE Journal of Selected Topics in Signal Processing},
	publisher    = {Institute of Electrical and Electronics Engineers (IEEE)},
	volume       = 17,
	number       = 1,
	pages        = {142–157},
	doi          = {10.1109/jstsp.2022.3224591},
	issn         = {1941-0484},
	url          = {http://dx.doi.org/10.1109/JSTSP.2022.3224591}
}

@misc{he2015deepresiduallearningimage,
	title        = {{Deep Residual Learning for Image Recognition}},
	author       = {Kaiming He and Xiangyu Zhang and Shaoqing Ren and Jian Sun},
	year         = 2015,
	url          = {https://arxiv.org/abs/1512.03385},
	eprint       = {1512.03385},
	archiveprefix = {arXiv},
	primaryclass = {cs.CV}
}

@misc{howard2019searchingmobilenetv3,
	title        = {{Searching for MobileNetV3}},
	author       = {Andrew Howard and Mark Sandler and Grace Chu and Liang-Chieh Chen and Bo Chen and Mingxing Tan and Weijun Wang and Yukun Zhu and Ruoming Pang and Vijay Vasudevan and Quoc V. Le and Hartwig Adam},
	year         = 2019,
	url          = {https://arxiv.org/abs/1905.02244},
	eprint       = {1905.02244},
	archiveprefix = {arXiv},
	primaryclass = {cs.CV}
}

@misc{irvin2019chexpertlargechestradiograph,
	title        = {{CheXpert: A Large Chest Radiograph Dataset with Uncertainty Labels and Expert Comparison}},
	author       = {Jeremy Irvin and Pranav Rajpurkar and Michael Ko and Yifan Yu and Silviana Ciurea-Ilcus and Chris Chute and Henrik Marklund and Behzad Haghgoo and Robyn Ball and Katie Shpanskaya and Jayne Seekins and David A. Mong and Safwan S. Halabi and Jesse K. Sandberg and Ricky Jones and David B. Larson and Curtis P. Langlotz and Bhavik N. Patel and Matthew P. Lungren and Andrew Y. Ng},
	year         = 2019,
	url          = {https://arxiv.org/abs/1901.07031},
	eprint       = {1901.07031},
	archiveprefix = {arXiv},
	primaryclass = {cs.CV}
}

@inproceedings{wang2017NIHchestxray,
	title        = {{ChestX-ray8: Hospital-scale Chest X-ray Database and Benchmarks on Weakly-Supervised Classification and Localization of Common Thorax Diseases}},
	author       = {Wang, Xiaosong and Peng, Yifan and Lu, Le and Lu, Zhiyong and Bagheri, Mohammadhadi and Summers, Ronald},
	year         = 2017,
	booktitle    = {IEEE Conference on Computer Vision and Pattern Recognition (CVPR)},
	pages        = {3462--3471},
    publisher    = {IEEE},
    address      = {New York, NY, USA},
}

@inproceedings{khan2024accuracy,
	title        = {Accuracy vs. Privacy: a Federated Learning Approach for Lung Ultrasound Pattern Classification},
	author       = {Khan, Umair and Custode, Leonardo Lucio and Smargiassi, Andrea and Inchingolo, Riccardo and Torri, Elena and Tursi, Francesco and Narvena, Veronica and Perrone, Tiziano and Demi, Libertario and Iacca, Giovanni},
	year         = 2024,
	booktitle    = {IEEE Ultrasonics, Ferroelectrics, and Frequency Control Joint Symposium (UFFC-JS)},
	pages        = {1--4},
	publisher    = {IEEE},
    address      = {New York, NY, USA},
}

@inproceedings{bilal2025enhancing,
	title        = {Enhancing Communication-Efficient Federated Learning for Human Activity Recognition Through Knowledge Distillation},
	author       = {Bilal, Hafiz Muhammad and Hassan, Mir and Iacca, Giovanni},
	year         = 2025,
	booktitle    = {IEEE Vehicular Technology Conference (VTC2025-Spring)},
	pages        = {1--7},
	publisher    = {IEEE},
    address      = {New York, NY, USA},
}

@inproceedings{han2025federated,
	title        = {Federated Learning for Lung Ultrasound Classification Across Age-diverse Patient Populations},
	author       = {Han, Xi and Khan, Umair and Smargiassi, Andrea and Inchingolo, Riccardo and Torri, Elena and Perrone, Tiziano and Zannin, Emanuela and Rigotti, Camilla and Cattaneo, Federico and Dognini, Giulia and others},
	year         = 2025,
	booktitle    = {IEEE International Ultrasonics Symposium (IUS)},
	pages        = {1--4},
	publisher    = {IEEE},
    address      = {New York, NY, USA},
}

@article{cansiz2026evolutionary,
	title        = {Evolutionary Algorithm-Based Ensemble of Aggregation Strategies in Federated Learning},
	author       = {Cansiz, Berke and Hassan, Mir and Taskiran, Murat and Iacca, Giovanni},
	year         = 2026,
	journal      = {IEEE Transactions on Evolutionary Computation},
	publisher    = {IEEE},
    address      = {New York, NY, USA},
    numpages     = 16
}

@INPROCEEDINGS{imagenet2009,
    author={Deng, Jia and Dong, Wei and Socher, Richard and Li, Li-Jia and Kai Li and Li Fei-Fei},
    booktitle={2009 IEEE Conference on Computer Vision and Pattern Recognition}, 
    title={ImageNet: A large-scale hierarchical image database}, 
    year={2009},
    volume={},
    number={},
    pages={248-255},
    doi={10.1109/CVPR.2009.5206848}
}

\appendix

\end{document}